\documentclass[a4paper]{article}

\usepackage{latexsym}
\usepackage{amsmath}

\long\def\nop#1{}

\def\comment{\edef\cps{\the\parskip} \parskip=0.5cm \begingroup \tt}

\expandafter\ifx\csname shortcite\endcsname\relax

\fi

\newbox\current

\long\def\plframebox#1{
\setbox\current\vbox{#1}		% set box

\vbox to \ht\current {\hrule\vss
\hbox to \wd\current {%
\vrule \hss\box\current\hss \vrule}
\vss\hrule }
}

\long\def\eatpar#1{%
\ifx#1\par                      % se il token e' \par
\let\nextmove=\eatpar           % rimetti \eatpar in coda
\else
\let\nextmove=#1%               altrimenti, rimetti il token in coda
\fi
\noexpand\nextmove%             il token o \eatpar viene rimesso in coda
}

\def\modifymargins#1#2{
\newdimen\addtoh
\newdimen\addtow
\addtoh=#1
\addtow=#2

\advance\topmargin by -\addtoh
\multiply\addtoh by 2
\advance\textheight by \addtoh

\advance\oddsidemargin by -\addtow
\advance\evensidemargin by -\addtow
\multiply\addtow by 2
\advance\textwidth by \addtow
}

\begingroup
\catcode`\~=11
\gdef\centertilde#1{\lower #1pt\hbox{~}}
\endgroup

\newcount\currenttime
\newcount\hour
\newcount\minute

\def\printtime{%
\currenttime=\time
\hour=\currenttime
\divide\hour by 60
\minute=-\hour
\multiply\minute by 60
\advance\minute by \currenttime
\the\hour:\ifnum\minute<10 0\fi\the\minute
}

\begingroup
\makeatletter
\global\let\@@date=\@date
\gdef\@date{\@@date\ --- \printtime}
\endgroup

\def\oggi{\number\day\space 
\ifcase\month\or
Gennaio\or Febbraio\or Marzo\or Aprile\or Maggio\or Giugno\or
Luglio\or Agosto\or Settembre\or Ottobre\or Novembre\or Dicembre\fi
\space \number\year}

\newcounter{rmexample}

\def\proof{\noindent {\sl Proof.\ \ }}

\def\qed{\hfill{\boxit{}}
  \ifdim\lastskip<\medskipamount \removelastskip\penalty55\medskip\fi}
\def\qedn#1{\hfill{\boxit{}$_#1$}
  \ifdim\lastskip<\medskipamount \removelastskip\penalty55\medskip\fi}
\long\def\boxit#1{\vbox{\hrule\hbox{\vrule\kern3pt
                  \vbox{\kern3pt#1\kern3pt}\kern3pt\vrule}\hrule}}

\def\M{{\cal M}} \def\N{{\cal N}}

\def\ie{i.e.}
\def\eg{e.g.}

\def\M{{\cal M}}

\def\C{{\rm C}}

\def\p{{\rm P}}
\def\np{{\rm NP}}

\def\S#1{\mbox{$\Sigma^p_{#1}$}}
\def\P#1{\mbox{$\Pi^p_{#1}$}}

\def\pspace{{\rm PSPACE}}

\def\c{$\leadsto$}

\def\cC{\c{\rm C}}

\def\nuc#1{\mbox{$\parallel\!\leadsto$#1}}

\def\nucC{\nuc{\rm C}}
\def\nucp{\nuc{\rm P}}

\def\profont{\sf}

\def\x3c{{\profont x3c}}

\def\possnewtheorem#1#2{
\expandafter\ifx\csname #1\endcsname\relax
\newtheorem{#1}{#2}
\fi
}

\def\possnewtheoremthree#1[#2]#3{
\expandafter\ifx\csname #1\endcsname\relax
\newtheorem{#1}[#2]{#3}
\fi
}

\possnewtheorem{theorem}{Theorem}
\possnewtheorem{corollary}{Corollary}
\possnewtheorem{lemma}{Lemma}
\possnewtheoremthree{proposition}[theorem]{Proposition}
\possnewtheorem{definition}{Definition}
\possnewtheorem{question}{Question}
\possnewtheorem{example}{Example}
\possnewtheorem{nontheorem}{Counterexample}
\possnewtheorem{property}{Property}
\possnewtheorem{assumption}{Assumption}
\possnewtheorem{conjecture}{Conjecture}
\newenvironment{theorem*}[1]{{\noindent \bf Theorem~#1}\begin{it}}{\end{it}\

}

\title{On the size of data structures used in symbolic model checking}
\author{Andrea Ferrara \and Paolo Liberatore \and Marco Schaerf \\
Dipartimento di Informatica e Sistemistica\\
Sapienza Universit\`a di Roma\\
Via Ariosto 5, 00185 Roma, Italy\\
Email: {\tt lastname@dis.uniroma1.it}}

\begin{document}

\date{}
\maketitle
%%%%%%%%%%%%% begin:abstract %%%%%%%%%%%%%%
\begin{abstract}

Temporal Logic Model Checking is a verification method in which we describe a
system, the model, and then we verify whether some properties, expressed in a
temporal logic formula, hold in the system. It has many industrial
applications. In order to improve performance, some tools allow preprocessing
of the model, verifying on-line a set of properties reusing the same compiled
model; we prove that the complexity of the Model Checking problem, without any
preprocessing or preprocessing the model or the formula in a polynomial data
structure, is the same. As a result preprocessing does not always exponentially
improve performance.

Symbolic Model Checking algorithms work by manipulating sets of states, and
these sets are often represented by BDDs. It has been observed that the size of
BDDs may grow exponentially as the model and formula increase in size. As a
side result, we formally prove that a superpolynomial increase of the size of
these BDDs is unavoidable in the worst case. While this exponential growth has
been empirically observed, to the best of our knowledge it has never been
proved so far in general terms. This result not only holds for all types of
BDDs regardless of the variable ordering, but also for more powerful data
structures, such as BEDs, RBCs, MTBDDs, and ADDs.

\end{abstract}

%%%%%%%%%%%%% end:abstract %%%%%%%%%%%%%%
 %

%%%%%%%%%%%%% begin:introduction %%%%%%%%%%%%%%
\section{Introduction}

Temporal Logic Model Checking \cite{CGP:2000} is a verification method for
discrete systems. In a nutshell, the system, often called the model, is
described by the possible transitions of its components, while the properties
to verify are encoded in a temporal modal logic. It is used, for example, for
the verification of protocols and hardware circuits \cite{BBGGY:1994}. Many
tools, called model checkers, have been developed to this aim. The most famous
ones are SPIN \cite{Hol:1997} and SMV \cite{McM:1993} (with its many
incarnations: NuSMV \cite{CCG:2002}, RuleBase \cite{BBL:1998}), VIS
\cite{VIS:1996}, and FormalCheck \cite{FormalCheck:1996}.

There are many languages to express the model; the most widespread ones are
Promela and SMV. Two temporal logics are mainly used to define the
specification: CTL \cite{CGP:2000} and LTL \cite{Pnu:1977}. In this paper we
focus on the latter.

In many cases, the two inputs of the model checking problem (the model and the
formula) can be processed in a different way. If we want to verify several
properties of the same system, it makes sense to spend more time on the model
alone, if the verification of the properties becomes faster. Many tools allow
to build the model separately from checking the formula
\cite{NuSMV:2007,VIS:2003,RULEBASE:2003}. This way, one can reuse the same
model, compiled into a data structure, in order to check several formulae.

In the same way, we may wish to verify the same property on different systems:
the property is this time the part we can spend more time on. Many tools allow
populating a property database \cite{NuSMV:2007,VIS:2003,RULEBASE:2003}, i.e.,
a collection of temporal formulae which will be checked on the models. We
imagine a situation in which we early establish the requirements that our
system must satisfy, even before the system is actually designed. As a result,
and we can fill a database of temporal formulae, but we do not yet describe
the system. While the design/modeling of the system goes on, we can preprocess
the formulae (without knowledge of the model, which is not yet known).
Whenever the system is specified, we can then use the result of this
preprocessing step to check the model against the formulae.

In this paper, we analyze whether preprocessing a part of the model checking
problem instances improve the performances. The technical tool we use is the
compilability theory \cite{CDLS:02,Lib:01}. This theory characterizes the
complexity of problems when the problem instances can be divided into two
parts (the fixed and the varying part), and we can spend more time on the
first part alone, provided that the result of this preprocessing step has
polynomial size respect the fixed part. We show that the Model Checking
problem remains PSPACE-hard even if we can preprocess either the model or the
formula, if this preprocessing step is constrained to have a polynomial size.
These theorems hold for all model checkers.

Finally, we answer to a long-time standing question in Symbolic Model Checking
\cite{McM:1993,DBLP:journals/iandc/BurchCMDH92}. It has been observed that the
BDDs that are used by SMV and other Symbolic Model Checking systems become
exponentially large in some cases. However, it has not yet been established
whether this size increase is due to the choice of variable ordering, or to
the kind of BDDs employed, or it is intrinsic of the problem. We show that, if
$\pspace \not\subseteq \P{2} \cap \S{2}$, such a growth is, in the worst case,
unavoidable. This result is independent from the particular class of BDDs and
from the variable order of the BDDs. It also holds for all decision diagrams
representing integer-value functions whose evaluation problem is in the
polynomial hierarchy, such as BEDs \cite{WBCG:2000}, BMD and *BMD
\cite{DBLP:journals/sttt/BryantC01}, RBCs \cite{ABE:2000}, MTBDDs
\cite{CFMMYZ:1993}, and ADDs \cite{BFGHMS:1993}.

%%%%%%%%%%%%% end:introduction %%%%%%%%%%%%%%
 %

%%%%%%%%%%%%% begin:preliminaries %%%%%%%%%%%%%%
\section{Preliminaries}

%%%%%%%%%%%%% begin:ltl %%%%%%%%%%%%%%
\subsection{Model Checking}

In this section, we briefly recall the basic definitions
about model checking that are needed in the rest of the
paper. We follow the notation of \cite{SC:1985,Schn:2002}.
LTL (Linear Temporal Logic) is a modal logic aimed at
encoding how states evolve over time. It has three unary
modal operators ($X$, $G$, and $F$) and one binary modal
operator ($U$). Their meaning is: $X\phi$ is true in
particular state if and only if the formula $\phi$ is true
in the next state; $G\phi$ is true if and only $\phi$ is
true from now on; $F\phi$ is true if $\phi$ will become true
at some time in the future; $\phi U \psi$ is true if $\psi$
will eventually become true and $\phi$ stays true until
then. We indicate with $L(O_1, \dots, O_n)$ the LTL fragment
in which the only temporal operators allowed are $O_1,
\dots, O_n$; for instance, $L(F,X)$ is the fragment of LTL in
which only $F$ and $X$ are allowed.

The semantics of LTL is based on Kripke models. 
In the following, for an 'atomic proposition' we mean a Boolean variable. 
Given a set of atomic proposition, a Kripke structure for LTL is a
tuple $\langle Q,R,\ell,I\rangle$, where $Q$ is a set of
states, $R$ is a binary relation over states (the transition
relation), $\ell$ is a function from states to atomic
propositions (it labels every state with the atomic
propositions that are true in that state), $I$ is a set of
initial states. A run of a Kripke structure is a Kripke
model. A Kripke model for LTL is an infinite sequence of
states, where the transition relation links each state
with the one immediately following it in the sequence. The
semantics of the modal operators is defined in the intuitive
way: for example, $F \phi$ is true in a state of a Kripke model 
if $\phi$ is true in some following state.

The main problem of interest in practice is to verify
whether all runs of a Kripke structure (all of its Kripke
models) satisfy the formula; this is the Universal Model Checking problem.
The Existential Model Checking one is to verify whether 
there is a run of the Kripke structure that satisfies the formula.  
In formal verification, we encode the behavior of a system as a Kripke structure, and the 
property we want to check as an LTL formula. Checking the
structure against the formula tells whether the system
satisfies the property. Since the Kripke structure is usually
called a ``model'' (which is in fact very different from a
Kripke model, which is only a possible run), this problem is
called Model Checking.

In practice, all model checkers describe a system by the Kripke structure
of its components. A Kripke structure can be seen as a transition system \cite{CGP:2000}.
Thus the global system is obtained by parallel composition of the transition systems
representing its components and sharing some variables \cite{MP:1995,CGP:2000}; 
using this approach, we can give results valid
for all model checkers. 

\iffalse
For example, a domain in which the states are represented by
the values of 100 Boolean variables which may change value
at any time step independently, then we have $2^{100}$
states, and the transition relation is composed of $2^{100}
\cdot 2^{100}$ pairs of states.
\fi

%%%%%%%%%%%%% end:ltl %%%%%%%%%%%%%%
 %

%%%%%%%%%%%%% begin:concurrent %%%%%%%%%%%%%%
\subsection{Composition of Transition Systems}

\iffalse
A model (a Kripke structure) is the representation of a
system. Often, the various parts composing the system
can be modeled separately, with a number of variables that are
shared between the description of the various parts.
\fi

Each component of the global system is modeled using a transition
system, which is a formal way to describe a possible
transition a system can go through. Intuitively, all is
needed is to specify the state variables, the possible
initial states, and which transitions are possible, \ie, we
have to say whether the transition from state $s$ to state
$s'$ is possible for any pair of states $s$ and $s'$. The
formal definition is as follows \cite{MP:1995,CGP:2000}.

\begin{definition}

A \textit{finite-state transition system} is a triple
$(V,I,\varrho)$, where $V=\{x_1,\ldots,x_n\}$ is a set of
Boolean variables, $I$ is a formula over $V$, and
$\varrho(V, V')$ is a formula over $V \cup V'$, where
$V'=\{x_1',\ldots,x_n'\}$ is a set of new variables 
in one to one relation with elememts of $V$.

\end{definition}

Intuitively, $V$ is the set of state variables, $I$ is a
formula that is true on a truth assignment if and only if it
represents a possible initial state, and $\varrho$ is true
on a pair of truth assignments if they represent a possible
transition of the system. The set of variables $V'$ is
needed because $\varrho$ must refer to both the value of a
variable in the current state ($x_i$) and in the next state
($x_i'$). In other words, in this formula $x_i$ means the
value of $x_i$ in the current state, while $x_i'$ is the
value of the same variable in the next state. For example,
the fact that $x_i$ remains true is encoded by
$\varrho=x_i \rightarrow x_i'$: if $x_i$ is true now, then
$x_i'$ is true, \ie, $x_i$ is true in the next state. 

Formally, a \textit{state} $s$ is an assignment to the
variables; a state $s'$ is \textit{successor} of a state $s$
iff $\langle s,s'\rangle \models \varrho(V,V')$. A
\textit{computation} is an infinite sequence of states $s_0,
s_1, s_2, \dots,$ satisfying the following requirements:

\begin{description}

\item[Initiality:] $s_0$ is initial, i.e. $s_0 \models I$

\item[Consecution:] For each $j \geq 0$, the state $s_{j+1}$
is a successor of the state $s_j$

\end{description}

For the sake of simplicity, without loss of any generality,
we only consider Boolean variables and Boolean assertions.
\iffalse
In fact, any assertion on enumerative variables is
polynomially reducible to a Boolean assertion on Boolean
variables.

Moreover the results given are independent from
fairness and from observability of variables and then we
ignore the fairness requirements and any consideration about
observability of variables which instead appears in
\cite{MP:1995}.
\fi

In order to model a complex system, we assume that each of
its parts can be modeled by a transition system. Clearly,
there is usually some interaction between the parts; as a
result, some variables may be {\em shared} between the
transition systems. In the following, we consider $k$
transition systems $M_1, \dots, M_k$.  Every $M_i$ is
described by $((V_i^L \cup V_i^S), I_i(V_i),
\varrho_i(V_i,V'_i))$ for i $1 \leq i \leq k$ where $V_i^L$
is the set variables local to $M_i$, $V_i^S$ is the set of
shared variables of $M_i$, and $V_i=V_i^L \cup V_i^S$. A
group of transition systems can be composed in different
ways: synchronous, interleaved asynchronous, and
asynchronous. The third way is not frequently used in Model
Checking, so we only define the first two ways of composition. 
In the following, a process is any of the
transition systems $M_i$. \\

The synchronous parallel composition of $k$ transition
systems is obtained by assuming that the global transition
is due to all processes $M_i$ making a transition
simultaneously. In other words, all processes must make a
transition at any time step, and no process is allowed to
``idle'' at any time step.

\begin{definition}

The \textit{synchronous parallel composition} of processes $M_1, \dots, M_k$,
denoted by $M_1 \| \dots \| M_k$, is the transition system $M=(V,I,\varrho)$
described by:

\[
\begin{array}{lll}
\\
V
=
\bigcup^k_{i=1} V_i
&~&
I(V)
=
\bigwedge^k_{i=1} I_i(V_i)
\\
\\
\varrho(V,V')
=
\bigwedge^k_{i=1} \varrho_i(V_i,V'_i)
\\
\\
\end{array}
\]

\iffalse
\begin{eqnarray*}
V
&=&
\bigcup^k_{i=1} V_i
\\
I(V)
&=&
\bigwedge^k_{i=1} I_i(V_i)
\\
\varrho(V,V')
&=&
\bigwedge^k_{i=1} \varrho_i(V_i,V'_i)
\end{eqnarray*}
\fi

\end{definition}

The basic idea of the interleaved asynchronous parallel
composition is that only one process is active at the same
time. As a result, a global transition can only result from
the transition of a single process. The variables that are
not changed by this process must maintain the same value.

\begin{definition}

The \textit{interleaved asynchronous parallel composition}
of $M_1,\dots, M_k$ is the transition system
$M=(V,I,\varrho)$: 
, where $V$ and $I$ are as in the synchronous composition and $\varrho$ is:

\[
\begin{array}{lll}
\\
\varrho(V,V')
=
\bigvee^k_{i=1}
\left[
\varrho_i(V_i,V'_i)
\wedge
\bigwedge^k_{
{j=1}\atop{j \neq i}
}
V^L_i={V^L_i}'
\right]
\\
\\
\end{array}
\]

\iffalse
\[
\begin{array}{lll}
V
=
\bigcup^k_{i=1} V_i
&~&
I(V)
=
\bigwedge^k_{i=1} I_i(V_i)
\\
\\
\varrho(V,V')
=
\bigvee^k_{i=1}
\left[
\varrho_i(V_i,V'_i)
\wedge
\bigwedge^k_{
{j=1}\atop{j \neq i}
}
V^L_i={V^L_i}'
\right]
\end{array}
\]
\fi

\iffalse
\begin{eqnarray*}
V
&=&
\bigcup^k_{i=1} V_i
\\
I(V)
&=&
\bigwedge^k_{i=1} I_i(V_i)
\\
\fi

\iffalse
% ultimo cambiamento
\[
\varrho(V,V')
=
\bigvee^k_{i=1}
\left[
\varrho_i(V_i,V'_i)
\wedge
\bigwedge^k_{
{j=1}\atop{j \neq i}
}
V^L_i={V^L_i}'
\right]
\]
\fi

The interleaved asynchronous parallel composition of $M_1, \dots,
M_k$, is denoted by $M_1 | \dots| M_k$.
\end{definition}

A model can be described as the composition of transition
systems. As a result, we can define the model checking
problem for concurrent transition systems as the problem of
verifying whether the model described by the composition of
the transition systems satisfies the given formula.

%%%%%%%%%%%%% end:concurrent %%%%%%%%%%%%%%
 %

%%%%%%%%%%%%% begin:comp %%%%%%%%%%%%%%
\subsection{Complexity and Compilability}

\iffalse
We assume that the reader knows the basic concept of the
theory of \np-completeness \cite{stoc-76,gare-john-79}. 
\fi
We assume that the reader knows the basic concepts of complexity 
theory \cite{stoc-76,gare-john-79}.
What we mainly use in this paper are the concepts of polynomial
reduction and the class \pspace.

The Model Checking problem is \pspace-complete, and is thus
intractable. On the other hand, as said in the Introduction,
it makes sense to preprocess only one part of the problem
(either the model or the formula), if this reduces the
remaining running time. The analysis of how much can be
gained by such preprocessing, however, cannot be done using
the standard tools of the polynomial classes and reductions.
The compilability classes \cite{CDLS:02} have to be used
instead.

The way in which the complexity of the problem is identified
in the theory of \np-completeness is that of giving a set of
increasing classes of problems. 
If a problem is in a class
\C\  but is not in an inner class $\C'$, then we can say
that this problem is more complex to solve that a problem in
$\C'$. A similar characterization, with similar classes, can
be given when preprocessing is allowed. For example the
class \nucp\  is the class of problems that can be solved in
polynomial time after a preprocessing step. Crucial to this
definition are two points:

\begin{enumerate}
\itemsep=-3pt
\item which part of the problem instance
can be preprocessed?

\item how expensive is the preprocessing part
allowed to be?

\end{enumerate}

The first point depends on the specific problem and on the
specific settings: depending on the scenario, for example,
we can preprocess either the model or the formula for the
model checking problem. The second question instead allows
for a somehow more general answer. First, we cannot limit
this phase to take polynomial time, as otherwise there would
be no gain in doing preprocessing from the point of view of
computational complexity. Second, we cannot allow the final
result of this part to be exponentially large, 
for practical reasons; we bound the
result of the preprocessing phase only to take a polynomial
amount of space.

In order to denote problems in which only one part can be
preprocessed, we assume that their instances are composed of
two parts, and that the part that can be preprocessed is the
first one. As a result, the model checking problem
written as $\langle M, \phi \rangle$ indicates that $M$ can be
preprocessed; written as $\langle \phi, M \rangle$ indicates that
$\phi$ can be preprocessed.

The ``complexity when preprocessing is allowed'' 
is established by characterizing how hard a problem is {\em
after} the preprocessing step. This is done by building over
the usual complexity classes: if \C\  is a ``regular''
complexity class such as \np, then a problem is in the
(non-uniform) compilability class \nucC\  if the problem is in \C\  after a
preprocessing step whose result takes polynomial space. In
other words, \nucC\  is ``almost'' \C, but preprocessing is
allowed and will not be counted in the cost of solving the
problem. More details can be found in \cite{CDLS:02}.

In order to identify how hard a problem is, we also need a
concept of hardness. Since the regular polynomial reductions
are not appropriate when preprocessing is allowed, ad-hoc
reductions (called nu-comp reductions in \cite{CDLS:02})
have been defined. 

\iffalse

As it was soon evident, the
reductions that the most natural for the compilability
classes are not very useful, as very few problem could be
proved hard using them. For this reason a new set of classes
\nucC\  have been defined. Proving a problem hard for \nucC\
implies that this problem is not in any class \cC' with $\C'
\subseteq \C$. Proving for example that a problem is
\nucC-hard implies that preprocessing does not make this
problem polynomial. As a result, proving hardness for the
new compilability class \nucC\  implies lower bound on the
complexity of the problems when polynomial-size-output
preprocessing is allowed.

\fi

In this paper, we do not show the hardness of problems
directly, but rather use a sufficient condition called
representative equivalence. For example, in order to prove
that model checking is \nuc\pspace-hard, we first show a
(regular) polynomial reduction from a \pspace-hard problem
to model checking and then show that this reduction
satisfies the condition of representative equivalence. 

Let us assume that we know that a given problem $A$ is
$\nucC$-hard and we have a polynomial reduction from
the problem $A$ to the problem $B$. Can we use this
reduction to prove the \nucC-hardness of $B$ ? 
Liberatore \cite{Lib:01} shows sufficient conditions that should
hold on $A$ as well as on the reduction. 
If all these conditions are verified, then
there is a nucomp reduction from $*A$ to $B$, 
where $*A=\{\langle x, y \rangle\ | y \in A \}$,
thus proving the \nucC-hardness of $B$.

\begin{definition}[Classification Function]

A {\em classification function} for a problem $A$ is a
polynomial function $Class$ from instances of $A$ to
nonnegative integers, such that $Class(y) \leq ||y||$.

\end{definition}

\begin{definition}[Representative Function]

A {\rm representative function} for a problem $A$ is a
polynomial function $Repr$ from nonnegative integers to
instances of $A$, such that $Class( Repr( n) )=n$, and
that $||Repr(n)||$ is bounded by some polynomial in $n$.

\end{definition}

\begin{definition}[Extension Function]

An {\em extension function} for a problem $A$ is a polynomial function from
instances of $A$ and nonnegative integers to instances of $A$ such that, for
any $y$ and $n \geq Class(y)$, the instance $y' = Exte(y,n)$ satisfies the
following conditions:\eatpar

\begin{enumerate}
\itemsep=-2pt
\item $y \in A$ if and only if $y' \in A$;
\item $Class(y')=n$.
\end{enumerate}

\end{definition}

Let us give some intuitions about these functions.
Usually, an instance of a problem is composed of a
set of objects combined in some way. For problems on
boolean formulas, we have a set of variables combined
to form a formula.  For graph problems, we have a set
of nodes, and the graph is indeed a set of edges,
which are pairs of nodes.
The classification function gives the number of objects in an
instance. The representative function thus gives an instance with the given
number of objects. This instance should be in some way ``symmetric'', in the
sense that its elements should be interchangeable (this is because the
representative function must be determined only from the number of objects).
Possible results of the representative function can be
the set of all clauses of three literals over a given
alphabet, the complete graph over a set of nodes, the graph with no edges,
etc.
Let for example $A$ be the problem of propositional satisfiability. We can take
$Class(F)$ as the number of variables in the formula $F$, while $Repr(n)$ can
be the set of all clauses of three literals over an alphabet of $n$ variables.
Finally, a possible extension function is obtained by adding tautological
clauses to an instance.
Note that these functions are related to the problem $A$ only, and do not
involve the specific problem $B$ we want to prove hard, neither the specific
reduction used. We now define a condition over the polytime
reduction from $A$ to $B$. Since $B$ is a problem of pairs,
we can define a reduction from $A$ to $B$ as a pair of
polynomial functions $\langle r,h \rangle$ such that $x \in A$
if and only if $\langle r(x),h(x) \rangle \in B$.

\begin{definition}[Representative Equivalence]

Given a problem $A$ (having the above three functions), a problem of pairs
$B$, and a polynomial reduction $\langle r,h \rangle$ from $A$ to $B$, the
condition of representative equivalence holds if, for any instance $y$ of $A$,
it holds:\eatpar

\[
\langle r(y),h(y) \rangle \in B \mbox{ ~~ iff ~~ } \langle
r(Repr(Class(y)),h(y) \rangle \in B
\]

\end{definition}

The condition of representative equivalence can be
proved to imply that the problem $B$ is \nucC-hard,
if $A$ is \C-hard \cite{Lib:01}. 
As an example, we show these three functions for the $PLANSAT^*_1$ problem. 
$PLANSAT_1^*$ is the following problem of planning: giving a STRIPS \cite{FN:1971} instance $y=\langle P,O,I,G\rangle$ in which the operators have an arbitrary number of preconditions and only one postcondition, 
is there a plan for $y$? $PLANSAT_1^*$ is \pspace-Complete \cite{Byl:1991}.  
Without loss of generality we consider $y=(P,O\cup o_0,I,G)$, 
where $o_0$ is a operator which is always usable (it has no preconditions) and 
does nothing (it has no postconditions). 
We use the following notation: $P=\{x_1, \dots, x_n\}$, $I$ is the set of conditions true in the initial state, $G=\langle \M, \N \rangle$. 
A state in STRIPS is a set of conditions.
In the following we indicate with $\phi_i^h$ the $h$th
positive precondition of the operator $o_i$, 
with $\phi_i$ all its the positive preconditions, 
with $\eta_i^h$ its $h$th negative precondition, 
and with $\eta_i$ all its negative preconditions; 
$\alpha_i$ is the positive postcondition of the
operator $o_i$, $\beta_i$ is the negative postcondition of
the operator $o_i$. Since any operator has only one
postcondition, for every operator $i$ 
it holds that $\| \alpha_i \cup \beta_i \|=1$.

Since we shall use them in the following, 
we define a classification function, 
a representative function and a extension function for $PLANSAT^*_1$: 
\begin{description}
	\item[] \textit{Classification Function:} 
	  $Class(y)=\|P\|.$ 
	  Clearly, it satisfies the condition $Class(y)\leq \|y\|$.
	\item[] \textit{Representative Function:} 
	  $Repr(n)=\langle P_n,\emptyset,\emptyset,\emptyset\rangle $, 
	  where $P_n=\{x_1, \dots, x_n \}$. Clearly, this function is polynomial and satisfies the following conditions: 
	  (i) Class(Repr(n))=n, (ii) $\|Repr(n)\|\leq p(n)$ where p(n) is a polynomial. 
\iffalse	  
		\begin{itemize}
			\item [-] Class(Repr(n))=n
			\item [-] $\|Repr(n)\|\leq p(n)$ where p(n) is a polynomial
		\end{itemize}
\fi
	\item[] \textit{Extension Function:} 
	  Let $y=\langle P,O,I,G\rangle$ and $y'=Exte(y,n)=\langle P_n,O,I,G\rangle$.
%	  defined as follows: $Exte(y,n)=\langle P_n,O,I,G\rangle$. 
	  Clearly for any $y$ and $n$ s.t. $n \geq Class(y)$ $y'$ satisfies the following conditions: 
	  (i)$y\in A$ iff $y'\in A$, (ii) $Class(y')=n$.
\iffalse	  
		\begin{itemize}
			\item [-] $y\in A$ iff $y'\in A$
			\item [-] $Class(y')=n$
		\end{itemize}
\fi
\end{description}  

Given the limitation of space we
cannot give the full definitions for compilability, for
which the reader should refer 
to \cite{CDLS:02} for an introduction, 
to \cite{CDLS:99,CDLS:00} for an application to the succinctness of some formalisms, 
to \cite{Lib:01} for further applications and technical advances.

%%%%%%%%%%%%% end:comp %%%%%%%%%%%%%%
 %

%%%%%%%%%%%%% end:preliminaries %%%%%%%%%%%%%%
 %

%%%%%%%%%%%%% begin:results %%%%%%%%%%%%%%
\section{Results}

The Model Checking problem for concurrent transition systems
is \pspace-complete \cite{KVW:00}. 
%also for constant LTL formulae. 
In Section \ref{premc}, 
we prove that the following problems are \nuc\pspace-hard, which implies that
they remain
\pspace-hard even if preprocessing is allowed.

\begin{enumerate}

\item model checking on the synchronous and interleaved
asynchronous composition of transition systems, where the
transitions systems are the fixed part of the problem
and the LTL formula is the varying part;

\item the same problem, where the LTL formula is the fixed part 
and the transition system is the varying part;

\item given a set of transition 
systems and a formula as the fixed part, a state as the varying part,
checking whether the state is a legal initial state.

\end{enumerate}

We can conclude that preprocessing the model or the formula
does not lead to a polynomial algorithm for model checking.
We recall that the fixed part is preprocessed off-line in a
polynomial data structure during the preprocessing phase,
and the varying part is given on-line.

The relevance of the first two problems is clear: in
formal verification, it is often the case that many
properties (formulae) have to be verified over the same
system (the model, in this case modeled by the transition
systems); on the other hand, it may also be that the same
property has to be verified on different systems.

The result about the third problem is less interesting by
itself. On the other hand, we use it to prove that 
the superpolynomial growth of the size 
of the data structures (e.g. OBDDs) currently used 
in model checkers based on the
Symbolic Model Checking algorithms \cite{McM:1993} (such as
SMV and NuSMV) cannot be avoided in general. 
The result is independent from its variable ordering, 
and it holds for others data structures that can be employed. 
We show these results in Section \ref{bdd}.

\iffalse
The result hold for all data
structures representing a Boolean function (\eg,
propositional formulae, RBC, BED, etc.) such that evaluating
a propositional interpretation is a problem that is inside
the polynomial hierarchy. Instances of data structures
currently used in Symbolic Model Checking tools are BDDs,
Boolean Expression Diagrams (BEDs) \cite{WBCG:2000} and
Reduced Boolean Circuits (RBCs) \cite{ABE:2000}. Our results
hold also for data structures used to represent
integer-value functions, 
like Binary Moment Diagrams (BMD and *BMD \cite{DBLP:journals/sttt/BryantC01}), Multi terminal binary decision diagrams (MTBDDs) \cite{CFMMYZ:1993}, Algebraic Decision Diagrams
(ADDs) \cite{BFGHMS:1993}; see for details the survey
\cite{DS:2001}. 
\fi

We point out that most of Temporal Logic Model Checking
algorithms \cite{CGP:2000} fall in one of three classes:
Symbolic Model Checking algorithms, which work on symbolic
representation of $M$; algorithms based on Bounded Model
Checking \cite{BCCZ:1999} (i.e. based on reduction from
Model Checking into SAT); algorithms that work on an explicit
representation of $M$ (e.g. \cite{GPVW:1995}). 
Our results concerning the size of the BDD (or some other decision diagrams)
are valid for all algorithms of the first class. 

In the proofs of the following sections we consider Existential Model Checking problems,
but the results are valid also for the Universal case; 
in fact {\rm PSPACE} is closed under complementation also for compilability.

\subsection{Preprocessing Model Checking}
\label{premc}

We now identify the complexity of the Model Checking problem
when the preprocessing of the model (represented as the composition
of transition systems) is allowed,
%, in the interleaved case.With a similar proof, the result holds also for the synchronous case.
both in the synchronous and in the interleaved case.

\begin{theorem} 
\label{MCSYN}
The model checking problem for $k$ synchronous concurrent
process $MC_{syn}=\langle (M_1 || \dots || M_k), \varphi\rangle$
where $\varphi \in LTL$ is \nuc\pspace-hard, and remains 
\nuc\pspace-hard for $\varphi \in L(F,G,X)$.
\end{theorem}
\proof
It is similar to the proof of the Theorem \ref{MCASYN}. 
We carry out a reduction from the $PLANSAT^*_1$ problem, 
that satisfies the conditions of representative equivalence;
the main difference is about the LTL formula.
\qed

We now consider the Model Checking problem for concurrent
processes composed in a interleaved way when the model can
be preprocessed.

\begin{theorem}
\label{MCASYN}
The model checking problem for $k$ interleaved concurrent
process $MC_{asyn}=\langle (M_1 | \dots | M_k), \varphi\rangle$
where $\varphi \in LTL$ is \nuc\pspace -complete, and
remains \nuc\pspace-hard for $\varphi \in L(F,G,X)$.
\end{theorem}
\proof
We show a reduction, that translates an instance
$y \in PLANSAT^*_1$ into an instance
$\langle r(y),h(y)\rangle  \in M_{asyn}$, satisfying
the condition of representative equivalence. 
Given $y=\langle P,O,I,G\rangle  \in PLANSAT^*_1$
\begin{itemize}
		\item [-] $r(y)$ defines a concurrent transition systems $M_1, \dots, M_n$,
         where each $M_i$ is obtained from a variable $x_i \in P$ and it is described by:
			\begin{description}
				\item[] $V_i=\{x_i\}$ 
				\item[] $I_i(V_i)= (x_i) \vee (\neg x_i)$
				\item[] $\varrho_i(V_i,V'_i)= (x_i=0 \wedge x'_i=0) \vee (x_i=0 \wedge x'_i=1) \vee \\ 
				            (x_i=1 \wedge x'_i=0) \vee (x_i=1 \wedge x'_i=1)$
			\end{description}
			The process $M=M_1 \| \dots \| M_n$ represents all possible computations, starting from all possible initial 
			assignments, over the variables $x_1, \dots, x_n$. 
		\item [-] $h(y)=h(I,G,O)=\neg(\phi_I\wedge\phi_G\wedge\phi_O)$\\
	where:
		\begin{description}
			\item[] $\varphi_I=\underset{i\in I}{\bigwedge}x_i\wedge
			\underset{i\notin I}{\bigwedge}\neg x_i$
			\item[] $\varphi_G=F( \underset{i\in \M}{\bigwedge}x_i\wedge
			\underset{i\in \N}{\bigwedge}\neg x_i)$ 
			\item[] $\varphi_O=G\overset{m}{\underset{i=0}{\bigvee}}[
			                  \overset{\| \phi_i \|}{\underset{h=1}{\bigwedge}}\phi_i^h\wedge
			                  \overset{\| \eta_i \|}{\underset{h=1}{\bigwedge}}\neg\eta_i^h\wedge
			                  X\gamma_i
			                  \wedge
			                  \overset{n}{\underset{j=1}{\underset{j\neq i}{\bigwedge}}}(x_j \leftrightarrow Xx_j)
			                  ]$ \\
		 \newline	                  
		 where \\
		 \newline	                  
		       $ \gamma_i=\left\{ \begin{array}{rll}
		 														\alpha_i & \mbox{if} & \alpha_i\neq \emptyset \\						
		 														\neg \beta_i & \mbox{if} & \beta_i\neq \emptyset \\						
		 											 \end{array}\right.
		 	     $                 
		\end{description}
\end{itemize}
\begin{description}
	\item[] $\varphi_I$ adds constraints about the initial states of y represented by I.
	\item[] $\varphi_G$ adds constraints about the goal states of y represented by G: it tells that a goal state will be reached.
	\item[] $\varphi_O$ describes the operators in O: globally (i.e. in every state) one of the operators must be used to go in the next state; $\varphi_O$ also describes the nop operator $o_0$.
\end{description}

Now, we prove that  $y\in PLANSAT^*_1$ iff $\langle r(y),h(y)\rangle\in M_{asyn}$.
Given $y=\langle P,O,I,G\rangle$, a solution for $y$ is a plan which generates the following sequence of states: $(s_1, \dots, s_p)$ where $s_1$ is an initial state and $s_p$ is a goal state. This sequence of states is obtained applying a sequence of operators $(o_{h_1}, \dots, o_{h_p})$
chosen in $O=\{o_1, \dots, o_m\}$ in the following way: for all i s.t. $1\leq i\leq p$, preconditions for $o_{h_i}$ are included in the state $s_i$, and the state $s_{i+1}$ is obtained from the state $s_i$ modifying the postcondition associated with $o_{h_i}$. We remark that a state in STRIPS is the set of conditions. 

The model $M=r(y)=r(P)$ represents all possible traces starting from all possible initial configurations, over the variables $x_1, \dots, x_n$.  
Thus, in this case the Existential Model Checking problem $\langle M,\varphi \rangle$ reduces to the satisfiability problem for $\varphi$: we check whether ther exists a trace among all traces over the variables $x_1, \dots, x_n$ that satisfies the LTL formula $\varphi$.  
Therefore, we have to prove that $y\in A$ iff $\varphi=h(y)$ is satisfiable:\\

$\Rightarrow$.  Given a solution for $y\in A$, we identify a model for $\varphi=h(y)$; by construction such a model has:
\begin{itemize}
	\item [-] initial state $s_1^M$ s.t. $\ell(s_1) = I \cup \{ \neg x_i | x_i \notin I \}$
	\item [-] a state $s_p^M$ s.t. $\ell(s_p) \subseteq \M \cup \{\neg x_i | x_i \notin \N \}$
	\item [-] given a state $s_i^M$, $s_{i+1}^M$ is successor of $s_i^M$ iff
		\begin{itemize}
			\item [-] $\ell(s_i^M) \subseteq Precond(o_{h_i})$, 
			          where $Precond(o_{h_i})=\{x_j | x_j \in \phi_{h_i}\} \cup \{\neg x_j | x_j \in \eta_{h_i}\}$
			\item [-] $\ell(s_{i+1}^M)= \ell(s_i^M)\cup \alpha_i - \beta_i $ \\
			where $\alpha_i$ is the positive postcondition of $o_{h_i}$ and $\beta_i$ is the negative postcondition of 
			$o_{h_i}$.
		\end{itemize}
		\item [-] an infinite number of states: when the state $s_p$ is reached this state is repeated for at least once or for ever (applying the nop operator $o_0$), or it is possible, it depends from y, to apply any operators whose preconditions are satisfied by $\ell(s_p^M)$.
\end{itemize}
  
$\Leftarrow$. Let $(s_1^M, \ldots, s_p^M, \ldots)$ a model for $\varphi$, and 
let $s_p$ the goal state, that the first state satisfying $\varphi_G$. 
We obtain the sequence of states visited by a plan which is a solution for $y$, 
by cutting the states after the goal state $s_p$ and assigning $s_i = \ell(s_i^M)$; 
thus this sequence of states $(s_1, \ldots, s_p)$, 
associated with the plan, has by construction:
\begin{itemize}
	\item [-] initial state $s_1$ s.t. $s_1 = I \cup \{\neg x_i | x_i \notin I\}$
	\item [-] a state $s_p$ s.t. $s_p \subseteq \M \cup \{\neg x_i | x_i \notin \N\}$
	\item [-] given a state $s_i$, $s_{i+1}$ is successor of $s_i$ iff
		\begin{itemize}
			\item [-] $s_i \subseteq Precond(o_{h_i})$
			\item [-] $s_{i+1} = s_i \cup \alpha_i - \beta_i $ \\
			where $\alpha_i$ is the positive postcondition of $o_{h_i}$ and $\beta_i$ is the negative postcondition of 
			$o_{h_i}$.
		\end{itemize}
\end{itemize}
\qed

Now we show the complexity results, 
both in the synchronous and in the interleaved case, 
%in the interleaved case,  
when the formula can be preprocessed.
%With a similar proof the result holds also for the synchronous case.

\begin{theorem} \label{MC'SYN}
The model checking problem for k synchronous concurrent
process $MC'_{syn}=\langle \varphi, (M_1 || \dots || M_k) \rangle$
where $\varphi \in LTL$ is \nuc\pspace -complete, and
remains \nuc\pspace-hard for $\varphi \in L(F,G,X)$.
\end{theorem}
\proof
$PLANSAT_1^*$ is the following problem of planning: giving a STRIPS \cite{FN:1971} instance $y=\langle P,O,I,G\rangle$ in which the operators have an arbitrary number of preconditions and only one postcondition, 
is there a plan for $y$? $PLANSAT_1^*$ is \pspace-complete \cite{Byl:1991}.  
Without loss of generality we consider $y=(P,O\cup o_0,I,G)$, 
where $o_0$ is a operator which is always usable (it has no preconditions) and 
does nothing (it has no postconditions). 
We use the following notation: $P=\{x_1, \dots, x_n\}$, $I$ is the set of conditions true in the initial state, $G=\langle \M, \N \rangle$. A state in STRIPS is a set of conditions.

In the following we indicate with $\phi_i^h$ the $h$th
positive precondition of the operator $o_i$, and with
$\eta_i^h$ the $h$th negative precondition of the operator
$o_i$; $\alpha_i$ is the positive postcondition of the
operator $o_i$, $\beta_i$ is the negative postcondition of
the operator $o_i$. Since any operator has only one
postcondition, for every operator $i$ 
it hold that $\| \alpha_i \cup \beta_i \|=1$.
 
We show a polynomial reduction from the problem $A$ to the
problem $B$ that satisfies the condition of representative
equivalence. This proves
that $B$ is \nucC-hard, if $A$ is \C-hard; to apply this
condition we must define a \textit{Classification Function},
a \textit{Representative Function} and a \textit {Extension
Function} for $A$. Thus we use such a proof schema: we
define a \textit{Classification Function}, a
\textit{Representative Function} and a \textit{Extension
Function}  for $PLANSAT_1^*$, then
we show a polynomial reduction from an instance $y \in
PLANSAT_1^*$ to an instance $\langle r(y),h(y)\rangle  \in
MC'_{SYN}$ that satisfies the condition of representative
equivalence.

	Let $y=\langle P,O,I,G\rangle  \in PLANSAT_1^*$. We
define $r$ and $h$ as follows:

\begin{itemize}

\item [-]
$
r(y)=
r(P)=
\neg
\left\{
F(x_g)\wedge
G
\bigwedge^n_{i=0}
\left[
\neg(x_i \leftrightarrow Xx_i) \rightarrow
\bigwedge^n_{{j=1}\atop{j\neq i}}(x_j \leftrightarrow Xx_j)
\right]
\right\}
$

\item [-] $h(y)$ defines the transition systems $M_1 \|
\dots \| M_k$. The generic $M_i$ is obtained from the
operators $o_{i_1}, \dots, o_{i_{d_i}}$ whose postcondition
involves the variable $x_i \in P$; $d_i$ is the number of
such operators. We add the variable $x_g$; thus we have at
most as many processes as variables: if $k$ is the number of
variables used as postcondition of operators plus one, we
have $k \leq n+1$. Let $M_k$ the process associated with the
variable $x_g$; 
this variable is $0$ at the beginning and 
it becomes $1$ only when the goal of the $PLANSAT$ problem is reached. 
$M_i$, for i s.t. $1 \leq i < k$, is defined
by:

\begin{description}

\item[] $V_i=\bigcup_{q=1}^{d_i} \phi_{i_q} \cup \eta_{i_q}
\cup \alpha_{i_q} \cup \beta_{i_q}$

\item[] $I_i(V_i)= \underset{x_j \in I \cap V_i}{\bigwedge} x_j \wedge 
\underset{x_j \in \overline{I} \cup V_i}{\bigwedge} \neg x_j$

\item[] $\varrho_i(V_i,V'_i)= \bigvee_{k=1}^{d_i} \overset{\| \phi_{i_k} \|}{\underset{h=1}{\bigwedge}}\phi_{i_k}^h \wedge
\overset{\| \eta_{i_k} \|}{\underset{h=1}{\bigwedge}}\neg \eta_{i_k}^h \wedge
\neg(\underset{i\in \M}{\bigwedge}x_i\wedge
			\underset{i\in \N}{\bigwedge}\neg x_i) \wedge
(x'_i \equiv b_{i_k})$
			
where $ b_{i_k}=\left\{ \begin{array}{rll}
			1 & \mbox{if} & \alpha_{i_k}\neq \emptyset \\
 			0 & \mbox{if} & \beta_{i_k}\neq \emptyset \\
		 \end{array}\right.
            $ \\
			\end{description}

The process $M_k$ is defined by:

\begin{description}

\item[] $V_k=\{x_g\}$ 

\item[] $I_k(V)= (x_g=0)$

\item[] $\varrho_k(V_k,V'_k) =
\bigwedge_{i \in \M} x_i\wedge
\bigwedge_{i \in \N}\neg x_i \wedge x'_g=1$

\end{description}

\end{itemize}
	
Now we prove that this reduction is correct, i.e. $y \in
PLANSAT_1^*$ iff $\langle r(y),h(y)\rangle \in MC'_{SYN}$.

$\Rightarrow$.  Given a solution for $y \in PLANSAT_1^*$, we
show a path of M which satisfies $\varphi$ (r(y) defined above).

A solution for $y$ is a plan which generates the following sequence of states: $(s_1, \dots, s_p)$ where $s_1$ is a initial state and $s_p$ is a goal state.  This sequence of states is obtained by applying a sequence of operators $(o_{h_1}, \dots, o_{h_p})$.

By construction $M$ admits a path $(s_1^M, \dots, s_p^M,
s_{p+1}^M, \dots)$ s.t.:

\begin{itemize}
	\item[-] $\ell(s_i^M)= s_i \cup \neg x_g$ for i $1\leq i\leq p$
	\item[-] $\ell(s_{p+1}^M)= s_p \cup x_g$
\end{itemize}
This path satisfies $\varphi$:
\begin{itemize}
	\item [-] $\varphi$ does not constrain about the initial state, therefore every initial state of the model is legal;
	\item [-] $x_g \subseteq \ell(s_{p+1}^M)$, therefore $F(x_g)$ is true;
	\item [-] the path shown is s.t. only one variable change at a time, therefore the subformula under the Globally is true.
\end{itemize}

$\Leftarrow$. Given a path of $M$ which satisfies $\varphi$, we show a solution for $y\in PLANSAT_1^*$. \\
The path is a sequence $(s_1^M, \dots, s_p^M, s_{p+1}^M, \dots)$.  We can obtain the sequence of states visited by a plan for y in this way:
\begin{itemize}
	\item[-] $s_i =\ell(s_i^M) - \{ \neg x_g \}$ for i $1\leq i\leq p$;
	\item[-] we ignore the rest of the path of $M$.
\end{itemize}
\qed

\begin{theorem}
The model checking problem for $k$ interleaved concurrent
process $MC'_{asyn}=\langle \varphi, (M_1 | \dots | M_k) \rangle$
where $\varphi \in LTL$ is \nuc\pspace -complete, and
remains \nuc\pspace-hard for $\varphi \in L(F)$.
\end{theorem}
\proof 
We carry out a reduction from the $PLANSAT^*_1$ problem, 
that satisfies the conditions of representative equivalence. 
The proof is similar to the proof of the Theorem \ref{MC'SYN}.
\qed

Now we introduce the decision problem
$MC_{s_0}=\langle [M,\varphi],
s_{0}\rangle$, where $M$ is specified by the
interleaved parallel composition of $k$ transition systems
$M_1, \dots, M_k$, $\varphi \in L(F)$, and $s_{0}$ is
a specific state. $MC_{s_0}$ is true if the model
checking problem for concurrent transition system $\langle
M,\varphi\rangle$ has solution and $s_{0}$ is a
legal initial state \ie, is an initial state belonging to
$M$ that satisfies $\varphi$. 

\begin{theorem} \label{MCInitial}
$MC_{s_0}$ is \nuc\pspace-complete.
\end{theorem}
\proof
The hardness follows from a polynomial time reduction from the problem $\langle (P,O,G), I \rangle$, 
that can be easily shown \nuc\pspace-complete on the basis of the results in \cite{libe-04}. \\

We sketch the reduction. We encode each operator in $O$ into each process $M_i$, 
and the goal $G$ into the formula $\varphi$. 
We encode the set of initial states $I$ using $s_0$. 
\qed

%This theorem is not important itself, but we use it 
%in the following section.

\subsection{The Size of BDDs}
\label{bdd}

In this section we prove that the size of BDDs and others data structures 
increases superpolynomially with the
size of the input data, in the worst case, 
when are used in a Symbolic Model Checking algorithm. 

Let $M$ a model specified by $k$ concurrent transition
systems $M_1, \dots, M_k$, and let $\varphi$ an LTL (or a CTL or CTL*) formula.

\begin{theorem} \label{BDD}
If $\pspace \not\subseteq \P{2} \cap \S{2}$, then
there is not always a BDD of any kind and with any variable
order that is polynomially large and represents the set of
initial states consistent with $M$ and $\varphi$.
\end{theorem}

\proof The evaluation problem for any kind of BDD, 
i.e. giving a BDD and an assignment of its variables evaluate the BDD,
\iffalse 
$\langle f_{BDD}, s_{assign}\rangle $, where $f_{BDD}$ is a BDD and
$s_{assign}$ is an assignment for $f_{BDD}$
\fi
is in \p\ . If there exists a poly-size BDD
representing the set of initial states consistent with $M$ and $\varphi$, then we can compile $M$ and $\varphi$ in the BDD and evaluate the assignment (representing a initial state) in polynomial time. 
This implies that $MC_{s_0}$ is in \nucp. 
\iffalse
The assignment $s_{assign}$ represents an
initial state.
\fi
We know from Theorem \ref{MCInitial} that $MC_{s_0}$ is \nuc\pspace-complete.
Therefore if such a BDD exists, then \nuc \pspace $=$\nucp. 
Now, by applying Theorem 2.12 in \cite{CDLS:02}, we conclude that there is no 
poly-size reduction from $MC_{s_0}$ to
the evaluation problem for a BDD, if $\pspace \not\subseteq
\P{2} \cap \S{2}$. \qed

Symbolic Model Checking algorithms work by building a
representation of the set of the initial states of $M$ that
satisfy $\varphi$. In particular, this set is represented by
BDDs. Therefore, the last theorem proves that these
algorithms, in the worst case, end up with a BDD of
superpolynomial size. This result does not depend on the
kind of BDD used (free, ordered, etc.) and on the variable
ordering. 
On the contrary, it holds also when the states are labeled with enumerative variable;
in other words it holds not only for BDD but also for any decision diagram, 
provided that the evaluation problem over this representation of the states is in
a class of the polynomial hierarchy. 
% Instances of these representations are BEDs, RBCs, MTBDDs, ADDs.
More formally, we consider an arbitrary representation of a
set of states. The evaluation problem is that of determining
whether a state belongs to a set.

\begin{theorem}
Given a method for representing a set of states whose
evaluation problem is in a class \S{i} of the polynomial hierarchy, 
it is not always possible to represent in polynomial
space the set of legal initial states of a model $M$ 
% specified by $k$ concurrent transition systems $M_1, \dots, M_k$ and an LTL (or a CTL or CTL*) 
and a formula $\varphi$, provided that
$\S{i+1} \not= \P{i+1}$.
\end{theorem}

The proof of this theorem has the same structure of the proof of the Theorem \ref{BDD}.

\iffalse
The result holds for all data
structures representing a Boolean function (\eg,
propositional formulae, RBC, BED, etc.) such that evaluating
a propositional interpretation is a problem that is inside
the polynomial hierarchy. 
\fi
Instances of such data structures,
currently used in Symbolic Model Checking tools, are BDDs,
Boolean Expression Diagrams (BEDs) \cite{WBCG:2000} and
Reduced Boolean Circuits (RBCs) \cite{ABE:2000}. 
Our results hold 
also for data structures used to represent
integer-value functions, 
like Multi terminal binary decision diagrams (MTBDDs) \cite{CFMMYZ:1993}, Algebraic Decision Diagrams
(ADDs) \cite{BFGHMS:1993}; see for details the survey
\cite{DS:2001}. 

On the other hand, it is also possible to prove that 
the above two theorems cannot be stated unconditionally: 
indeed if $\p=\pspace$, then there is a data structure 
of polynomial size allowing the representation of the set 
of initial states in such a way deciding whether a state is 
in this set can be decided in polynomial time. As a result, 
the non-conditioned version of the above two theorems implies 
a separation in the polynomial hierarchy.

%%%%%%%%%%%%% end:results %%%%%%%%%%%%%%
 %

% input{discussion}
%%%%%%%%%%%%% begin:related %%%%%%%%%%%%%%
\section{Related Works}

Some works in the literature are related to the results in this article:

\begin{enumerate}

\item the exponential growth of the BDD size respect to a particular problem
(e.g.\ integer multiplication \cite{Bry:1991}); some results concern the size
growth of other decision diagrams \cite{DS:2001} respect to particular
problems. While these results are not conditional to the collapse of the
polynomial hierarchy as the ones reported in this paper, they are also more
specific, as they concern only specific kinds of data structures (e.g. OBDDs)
respect to particular problems (e.g. integer multiplication).

\item the complexity of model checking:

		\begin{enumerate}

				\item the parametrized complexity
\cite{DF:1999} of a wide variety of model checking problems
\cite{DBLP:conf/stacs/DemriLS02}, analyzing the state explosion problem;

				\item it has been shown that
\cite{DBLP:conf/lpar/FerraraPV05}:

				\begin{enumerate}
					\item the complexity of model checking does not decrease under the ipotheses of some structural restrictions 	(e.g. 													treewidth) in the input.				
					\item despite a CNF formula of bounded treewidth can be represented by an OBDD of polynomial size, 
								the nice properties of treewidth-bounded CNF formulas are not preserved under existential quantification or unrolling,
								that is a basic operation of model checking algorithms. 
	  		\end{enumerate} 

				\item the compilability of the model checking
problem \cite{DBLP:conf/aiia/FerraraLS07}: it remains \pspace-complete even if
a part of the input, either the implicit model or the formula, is preprocessed
using any amount of time and storing the result of this prerpocesing step in a
polynomial-sized data structure. 

		\end{enumerate} 
 
\item the theoretical limitations of Symbolic Model Checking. The state
explosion problem can be partially explained by complexity theoretic results
\cite{DBLP:conf/dagstuhl/ClarkeGJLV01} ; in fact, problems (also whose inputs
are graphs), usually increase their worst case complexity when the input is
represented by BDD or other Boolean formalisms
\cite{DBLP:journals/cjtcs/FeigenbaumKVV99, DBLP:journals/iandc/GalperinW83,
DBLP:journals/iandc/PapadimitriouY86, DBLP:journals/ai/Balcazar96,
DBLP:conf/wg/LozanoB89, BLT92, DBLP:conf/aaecc/Toran88,
DBLP:journals/eccc/ECCC-TR95-048, DBLP:journals/ipl/Veith97,
DBLP:journals/iandc/Veith98, DBLP:journals/apal/GottlobLV99}. Moreover, a
classic information theoretic argument shows that only a small fraction of all
finite Kripke structures can be exponentially compressed \cite{LV93}.

\item succinctness of languages; for instance \cite{DBLP:conf/kr/Coste-MarquisLLM04}, in which succinctness of language for preferences are discussed, and \cite{DBLP:journals/jair/DarwicheM02} that presents results on the succinctness of several formalism, including BDDs and CNF.
\end{enumerate}
%%%%%%%%%%%%% end:related %%%%%%%%%%%%%%
 %

\bibliographystyle{plain}

\end{document}